\documentclass[10pt,twocolumn,letterpaper]{article}

\usepackage[pagenumbers]{cvpr} %

\usepackage{amsmath}
\usepackage{bm}

\definecolor{cvprblue}{rgb}{0.21,0.49,0.74}
\usepackage[pagebackref,breaklinks,colorlinks,allcolors=cvprblue]{hyperref}

\def\paperID{} %
\def\confName{CVPR}
\def\confYear{2026}

\def\paperName{3D-RE-GEN}
\title{\paperName : \\ 3D Reconstruction of Indoor Scenes with a Generative Framework
}

\author{Tobias Sautter\\
{\tt\small tobias.sautter@student.uni-tuebingen.de}
\and
Jan-Niklas Dihlmann\\
{\tt\small jan-niklas.dihlmann@uni-tuebingen.de}
\and
Hendrik Lensch\\
{\tt\small hendrik.lensch@uni-tuebingen.de}
}

\begin{document}
\twocolumn[{%
\renewcommand\twocolumn[1][]{#1}%
\maketitle
\begin{center}
    \centering
    \captionsetup{type=figure}
    \includegraphics[width=\textwidth]{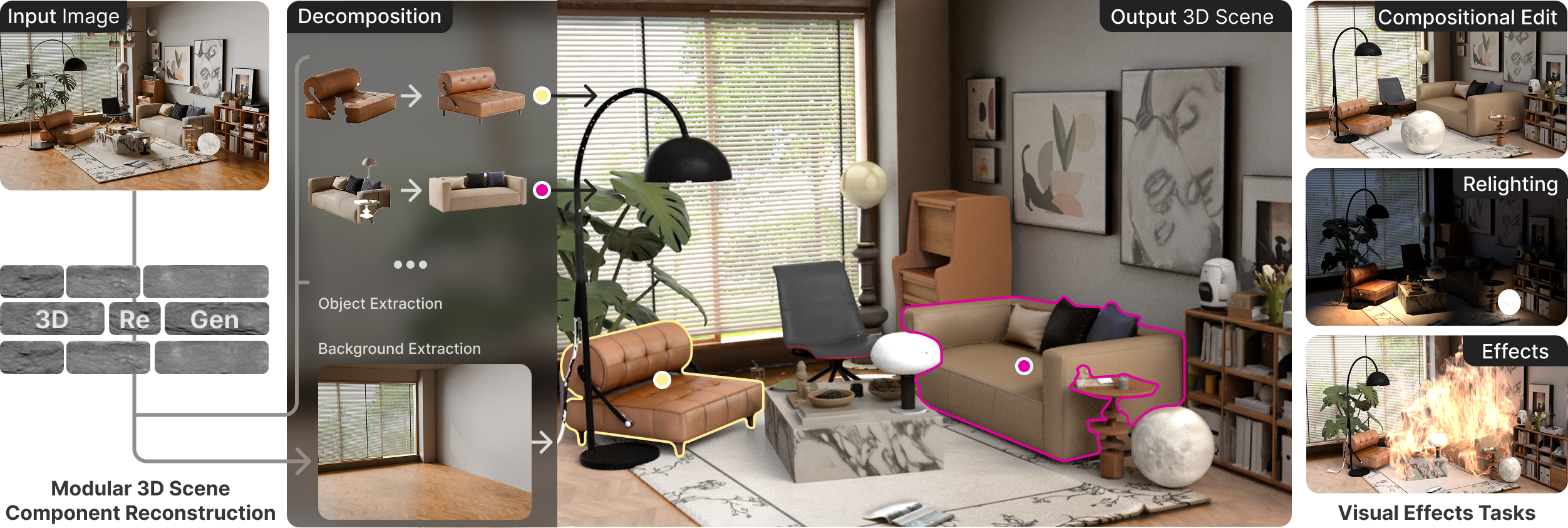}
    \captionof{figure}{\textbf{3D-RE-GEN in action}: A single image is decomposed into a clean background and a complete 3D scene with individual 3D objects, creating a production-ready asset for immediate use in VFX and games. Project page: \protect\url{https://3dregen.jdihlmann.com/}}
    \label{sec:title_image}
\end{center}%
}]

\begin{abstract}
Recent advances in 3D scene generation produce visually appealing output, but current representations hinder artists' workflows that require modifiable 3D textured mesh scenes for visual effects and game development. Despite significant advances, current textured mesh scene reconstruction methods are far from artist ready, suffering from incorrect object decomposition, inaccurate spatial relationships, and missing backgrounds. We present 3D-RE-GEN, a compositional framework that reconstructs a single image into textured 3D objects and a background. We show that combining state of the art models from specific domains achieves state of the art scene reconstruction performance, addressing artists' requirements. 

Our reconstruction pipeline integrates models for asset detection, reconstruction, and placement, pushing certain models beyond their originally intended domains. Obtaining occluded objects is treated as an image editing task with generative models to infer and reconstruct with scene level reasoning under consistent lighting and geometry. Unlike current methods, 3D-RE-GEN generates a comprehensive background that spatially constrains objects during optimization and provides a foundation for realistic lighting and simulation tasks in visual effects and games.
To obtain physically realistic layouts, we employ a novel 4-DoF differentiable optimization that aligns reconstructed objects with the estimated ground plane. 3D-RE-GEN~achieves state of the art performance in single image 3D scene reconstruction, producing coherent, modifiable scenes through compositional generation guided by precise camera recovery and spatial optimization. \protect\url{https://3dregen.jdihlmann.com/}

\end{abstract}
    
\section{Introduction}
\label{sec:intro}

The demand for immersive 3D content is a cornerstone of modern creative industries, most notably in Visual Effects (VFX) and game development. Traditionally, the creation of 3D scenes is a significant production bottleneck, requiring highly specialized artists with extensive experience. To construct a single scene, an environment artist must first model and texture the background. Then, every individual asset within that scene, from chairs to tables to small props, must be independently modeled and textured. Finally, a layout artist must meticulously assemble all these components into a coherent whole. Even for highly skilled professionals, this workflow is exceptionally time consuming and costly, scaling exponentially with scene complexity.
Recent generative models for single image to 3D (e.g., \cite{bossSF3DStableFast2024, zhaoHunyuan3D20Scaling2025}) offer the potential to accelerate asset creation and democratize 3D modeling. However, while these models are effective at generating isolated objects, they generally fail when applied to complex, multi object scenes.

Reconstructing a 3D scene from a single 2D image is an inherently ill-posed problem. A 2D projection lacks true depth information, and objects mutually occlude one another, resulting in incomplete structural data. Furthermore, accurately inferring the spatial positioning of objects to produce a credible layout remains a significant challenge, as the 2D image provides no explicit information about object specific distances, scale, or contact points.

Previous research has addressed this issue through various approaches. Early holistic methods attempted to reconstruct entire scenes in a single pass \cite{huangHolistic3DScene2018, zhangHolistic3DScene2021}, while retrieval based methods matched image objects to databases of 3D models \cite{gaoDiffCADWeaklySupervisedProbabilistic2024, izadiniaIM2CAD2017}. More recently, compositional and diffusion based methods have become prevalent \cite{huangMIDIMultiInstanceDiffusion2024, zhaoDepRDepthGuided2025, ardeleanGen3DSRGeneralizable3D2025, nieTotal3DUnderstandingJointLayout2020, mengSceneGenSingleImage3D2025}. Despite a strong emphasis on spatial relationships between objects, these methods share common critical gaps: the background environment is frequently neglected, and there is often no mechanism to enforce physical plausibility, such as aligning objects to a common ground plane.

This paper introduces \paperName, a framework designed to address this gap and generate production ready 3D scenes from a single image. The modular pipeline leverages large scale pretrained models and incorporates a novel four degree of freedom (4-DoF) constrained optimization with differentiable rendering. Our method estimates precise camera poses and generates an accurate textured background mesh; creates high quality 3D assets using a novel context aware inpainting technique; and aligns objects to the background through a robust optimization strategy, alternating between a five degree of freedom (5-DoF) model for free floating objects and the new 4-DoF model for ground aligned objects. 

Although \paperName~is developed with artists in mind, it demonstrates competitive performance and achieves state of the art results on comparative academic benchmarks. By automating the most laborious parts of scene reconstruction, \paperName~allows artists to move from a single concept image to a fully editable 3D environment in minutes, not days, drastically accelerating the creative workflow and minimizing needed experience. 

\section{Related Works}
\label{sec:related_works}

\begin{figure*}[t!]
    \centering
    \includegraphics[width=1\linewidth]{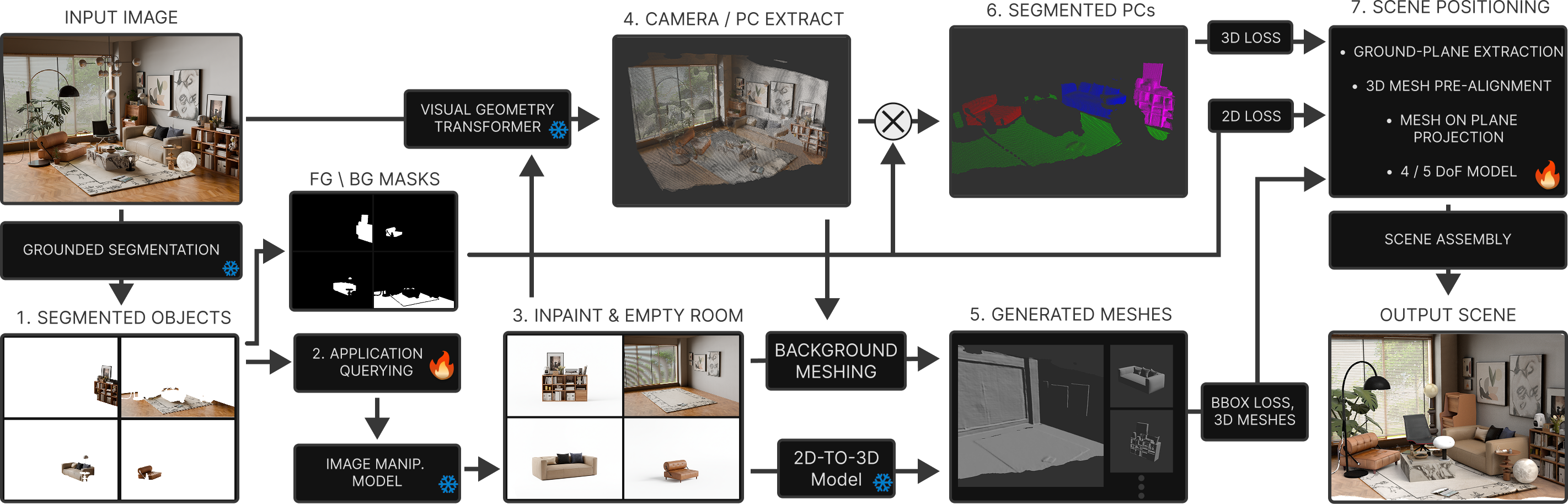}
    \caption{ \textbf{\paperName~Framework Overview.} Our framework converts a single image into a complete 3D scene. First, we segment the image: these masks provide the 2D silhouette loss, while our novel Application-Querying (A-Q) model generates clean, inpainted object images for 3D meshing. In parallel, the input and a generated "empty room" image are used to extract the camera and scene point cloud, which is masked to create the 3D geometric loss. Finally, our Scene Positioning model assembles the 3D assets and background by minimizing both losses, using a novel 4-DOF constrained workflow to ensure all ground based objects are physically aligned to the floor. }
    \label{fig:pipeline}
\end{figure*}

Reconstructing 3D scenes from single 2D images is a fundamental challenge in computer vision, as it requires synthesizing complex geometric structures and spatial relationships from limited visual data \cite{dahnertPanoptic3DScene2022, khanThreeDimensionalReconstructionSingle2022, zhangUni3DUnifiedBaseline2023, liuVoxelbased3DDetection2021, yaoFront2BackSingleView2020}.
In recent years, this task has attracted considerable research interest, with various approaches proposed to address the ambiguities and occlusions inherent in single-view observations \cite{liuHighFidelitySingleviewHolistic2022, gkioxariLearning3DObject2022, chuBUOLBottomUpFramework2024, popovCoReNetCoherent3D2020, wuAmodal3RAmodal3D2025a, xiePix2VoxContextaware3D2019, hanREPAROCompositional3D2025, huangMIDIMultiInstanceDiffusion2024}. 
The field has progressed from early methods that utilized explicit geometric priors to contemporary approaches that employ deep learning and generative models \cite{ardeleanGen3DSRGeneralizable3D2025, huangMIDIMultiInstanceDiffusion2024, wuUnique3DHighQualityEfficient2024}. 
This section reviews major developments, with particular emphasis on object level reconstruction \ref{sub:object_reconstruction}, scene composition strategies \ref{sub:scene_composition}, and diffusion-based techniques \ref{sub:diff_based_scene_gen_tech} that have emerged as promising solutions.

\subsection{Monocular Object-Level 3D Reconstruction}
\label{sub:object_reconstruction}
Recent advances in 3D representation learning have facilitated substantial progress in reconstructing individual objects from single images \cite{bossSF3DStableFast2024, laiHunyuan3D25HighFidelity2025, zhaoHunyuan3D20Scaling2025}. To extract the correct object from an image, models often leverage pretrained 2D object detectors or segmentation networks \cite{liSemanticSAMSegmentRecognize2023, renGroundedSAMAssembling2024}.
Early approaches primarily utilized supervised learning frameworks that mapped image features to 3D geometry, 
often employing explicit representations such as point clouds or meshes \cite{gkioxariLearning3DObject2022, huangHolistic3DScene2018}. 
These methods generally required extensive 3D supervision and exhibited limited generalization to novel object categories. 
The introduction of diffusion models has transformed the field, enabling the generation of high fidelity 3D shapes with enhanced generalization \cite{wuDirect3DScalableImageto3D2024, zhouZeroShotSceneReconstruction2024}. Diffusion based methods denoise latent representations of 3D geometry conditioned on image inputs, capturing complex shape priors without explicit 3D supervision. 
The availability of large scale 3D datasets has further accelerated progress, supporting the training of models capable of generating diverse and detailed 3D objects from single view observations \cite{fu3DFUTURE3DFurniture2020, fu3DFRONT3DFurnished2021, deitkeObjaverseUniverseAnnotated2022}.

\subsection{Scene Composition Strategies}
\label{sub:scene_composition}

Scene reconstruction introduces challenges beyond those encountered in object level reconstruction, 
as it necessitates modeling spatial relationships among multiple objects. Existing approaches are generally categorized by their strategies for composing individual objects into a coherent scene. 

Holistic methods process the entire scene as a single entity, predicting a unified 3D representation from the input image that includes all objects \cite{chuBUOLBottomUpFramework2024, dahnertPanoptic3DScene2022}. These methods often employ depth estimation as a geometric prior, projecting image features back into 3D space before utilizing encoder and decoder architectures to recover scene geometry. Although effective for simple scenes, holistic approaches often struggle with complex scenes containing multiple objects of varying scales and orientations, and they frequently yield low resolution reconstructions that do not generalize well to real world images \cite{chuBUOLBottomUpFramework2024, liuHighFidelitySingleviewHolistic2022}.

\subsection{Diffusion-based Scene Generation Techniques}
\label{sub:diff_based_scene_gen_tech}
Compositional approaches mitigate these limitations by decomposing scenes into individual objects, reconstructing each object independently, and subsequently assembling them into a unified scene. Early compositional methods utilized feed-forward networks to reconstruct objects from segmented image regions, followed by spatial alignment optimization \cite{nieTotal3DUnderstandingJointLayout2020, zhangHolistic3DScene2021}. 

More recent techniques leverage large scale, pretrained object reconstruction models to enhance the quality of individual reconstructions, followed by pose optimization to maintain spatial coherence \cite{ardeleanGen3DSRGeneralizable3D2025, hanREPAROCompositional3D2025, zhouZeroShotSceneReconstruction2024}. Despite these advances, compositional pipelines face two key challenges: (1) error accumulation across multiple stages, where mistakes in intermediate steps significantly affect the final outcome; and (2) the absence of global scene context during object reconstruction, which often results in misaligned spatial relationships among objects \cite{zhouZeroShotSceneReconstruction2024}.

Recent work extends single object diffusion models to multi object scenarios by modeling inter object relationships directly during generation \cite{huangMIDIMultiInstanceDiffusion2024, mengSceneGenSingleImage3D2025}. Depth-guided conditioning further enhances geometric consistency by integrating depth information throughout reconstruction \cite{zhaoDepRDepthGuided2025}. To handle occlusions, these state of the art methods often employ attention or feature aggregation modules to fuse partial object features with global scene context \cite{zhaoDepRDepthGuided2025, huangMIDIMultiInstanceDiffusion2024, mengSceneGenSingleImage3D2025}. However, these approaches still lack explicit mechanisms for robust scene level alignment and, critically, do not enforce strict physical alignment, which is essential for generating coherent scenes that preserve the input image’s spatial relationships. Our approach builds upon the compositional paradigm while addressing these limitations through novel mechanisms for scene alignment and object understanding. We introduce a novel Application-Querying method to achieve a global understanding of object properties, and a 4-DoF ground alignment constraint to overcome the spatial misalignment and physical implausibility issues inherent in existing methods. This combination enables a more accurate and coherent scene reconstruction that better matches the input image's visual content.

\section{Method}
\label{sec:method}

This chapter details the complete framework of our proposed 3D-RE-GEN pipeline. We present a step by step breakdown of our compositional diffusion approach, beginning with a single 2D input image and concluding with a fully reconstructed, physically plausible 3D scene. Figure \ref{fig:pipeline} shows a high level overview of this workflow. The entire pipeline commences with a single input image,~$\mathbf{I}$. From this image, individual objects are identified and segmented (Section \ref{section:image_extraction}). These segmented objects are then processed by our novel context aware inpainting method (Section \ref{section:novel_inpainting}), which serves two purposes: first, to generate clean, isolated images of each object on a white background, $\mathbf{I}_{obj\_fg}$, and second, to remove all objects from the original image, creating a clean background plate, $\mathbf{I}_{bg}$.

The newly generated, isolated object images are then passed to a 2D to 3D asset creation module (Section \ref{section:3d_priors}) to produce textured 3D meshes, $\mathcal{M}$. Based on the clean background, our scene reconstruction module (Section \ref{section:scene_recon}) places each asset back into the scene using our novel 4-DoF constrained optimization, ensuring a physically correct pose by aligning it with both the 3D scene geometry and the 2D object masks. The final output is a complete 3D scene, reconstructed from the initial 2D image, which is ready for direct use in applications such as game development, simulation, or visual effects.

\subsection{Image Extraction and Mask Refinement}
\label{section:image_extraction}

The input to our system is a single RGB image, $\mathbf{I}$. 
In the first step, Grounded SAM \cite{renGroundedSAMAssembling2024} performs text based instance segmentation to identify all objects of interest. 
A lightweight interactive tool enables further manual inspection and refinement of the generated masks, correcting any errors in the initial automated segmentation if needed. 
The quality of the resulting binary masks, $\mathbf{M}_{obj}$, is fundamentally important to the success of the entire pipeline. 
They provide the precise 2D silhouette ground truth for the $L_{silhouette}$ loss during the 3D pose refinement (Section \ref{section:scene_recon}) and they define the exact pixels to be extracted and provided to our inpainting model.

\subsection{Context Aware Object Inpainting}
\label{section:novel_inpainting}

A primary challenge in compositional scene reconstruction is handling occluded objects. Simply extracting a partial segment and tasking a generative model with its completion often fails, as the model lacks the necessary scene context to resolve ambiguities, a limitation noted in related generative assembly work~\cite{zhouZeroShotSceneReconstruction2024}. 
To address this, we introduce \textbf{Application Querying (A-Q)}, a novel visual prompting technique that provides a rich, structured representation of both the scene context and the generative task to a large scale, pretrained image editing model.

\begin{figure}[h]
    \centering

    \hfill
    \begin{subfigure}{0.99\linewidth}
        \centering
        \includegraphics[width=\linewidth]{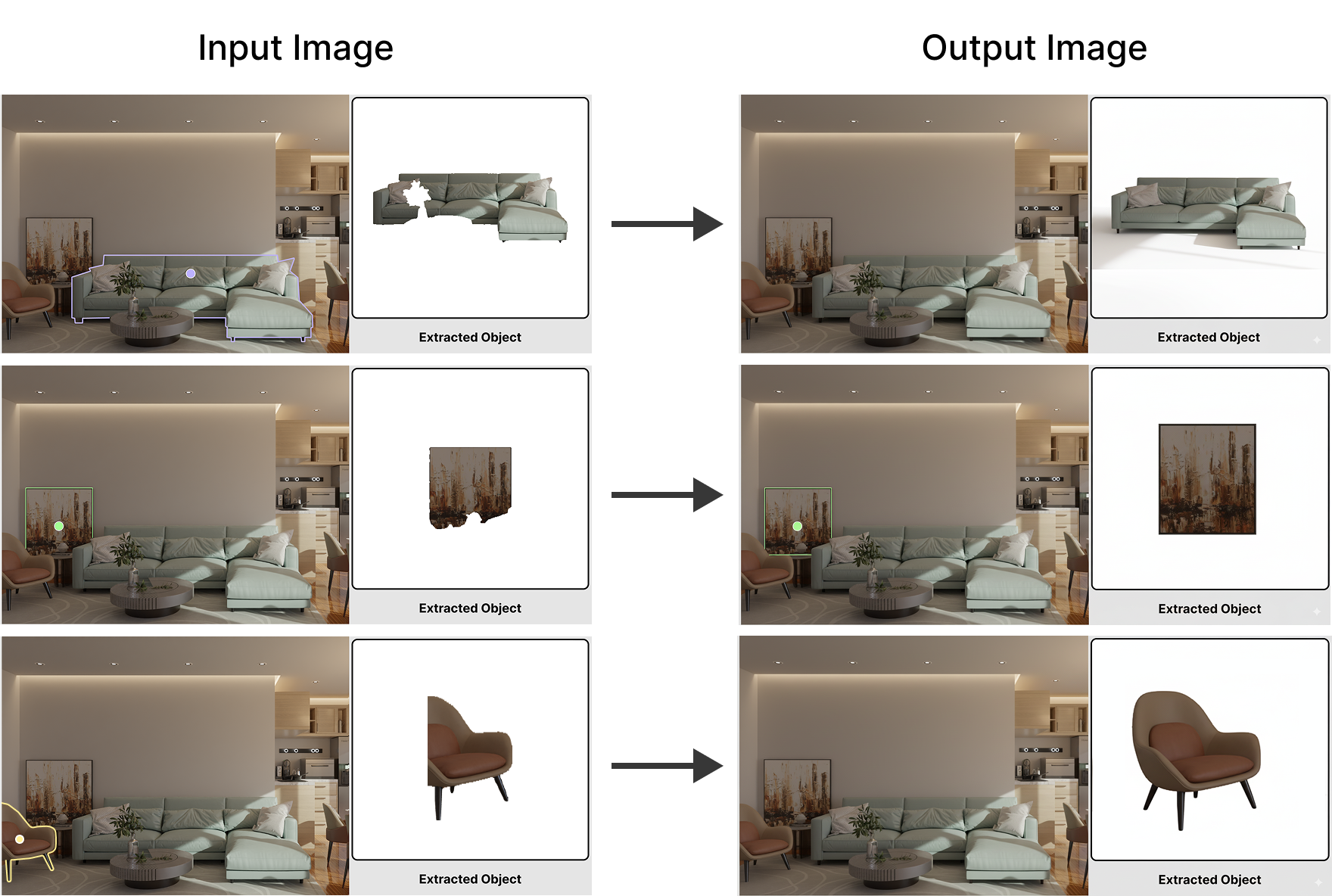}
        \caption{Input and output of the Application-Querying (A-Q) method.}
        \label{fig:in_and_out_AQ}
    \end{subfigure}

    \hfill
    \begin{subfigure}{0.99\linewidth}
        \centering
        \includegraphics[width=\linewidth]{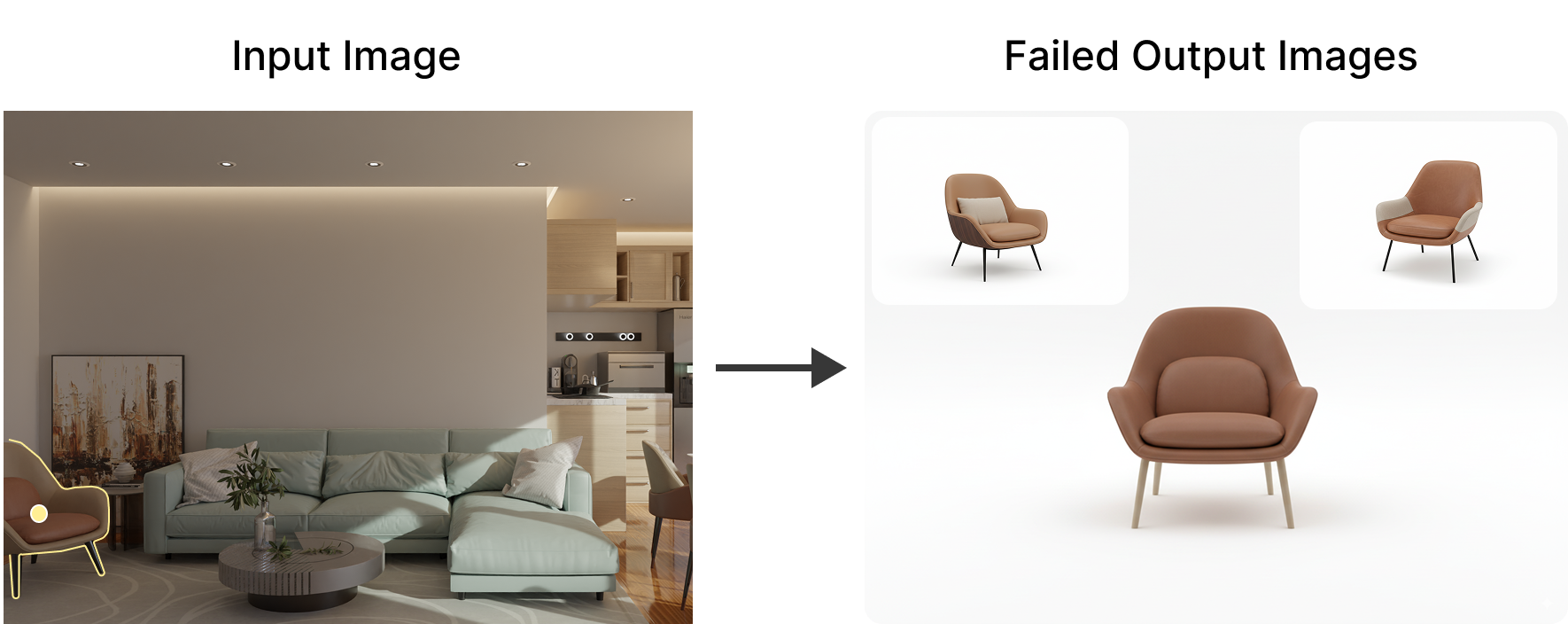}
        \caption{Failure cases. The object image completion won't work correctly without a proper UI image prompt, e.g., wrong materials and shapes.}
        \label{fig:in_and_out_AQ}
    \end{subfigure}    
    \caption{\textbf{Application-Querying.} Visual example of how we utilize a GUI-style interface to provide better scene context to the image manipulation model.}
    \label{fig:banana_seg}
\end{figure}

Rather than providing only the occluded object segment, our A-Q method constructs a composite query image. This image is structured to mimic a user interface, as shown in Figure \ref{fig:banana_seg}. The query presents the model with two distinct panels: one panel displays the full, original input image $\mathbf{I}$ with the target object's segmentation outline, providing complete scene context. The second panel displays the extracted, occluded object segment on a neutral white background, clearly defining the generative task.  Figure \ref{fig:in_and_out_AQ} is highlighting the importance of the correct UI image prompt for the model to correctly and consistently produce the correct result.

We then prompt the generative model to complete this visual query, tasking it with returning an image in the identical UI-style format, but with the object segment in the second panel fully inpainted. This structured query compels the model to leverage the contextual cues from the first panel, such as perspective, lighting, and surrounding style, to inform the generative completion of the object in the second panel. The model's output is then parsed to extract the completed object image, $\mathbf{I}_{obj\_fg}$. This resulting asset is an ideal input for 2D to 3D reconstruction models, as it combines a clean, isolated depiction with the high fidelity, context aware details inferred from the original scene.

\subsection{Generation of 3D Scene Geometry and Assets}
\label{section:3d_priors}

The core of our reconstruction pipeline relies on aligning generated 3D assets to the input image. To achieve this, we must first establish a set of 3D guidance points and constraints that define the scene's geometry and the camera's perspective. This process involves two parallel steps: (1) reconstructing the 3D scene geometry to derive camera pose and target point clouds, and (2) generating a 3D asset for each segmented object. \\
\textbf{Scene and Camera Reconstruction.}
To acquire a robust understanding of the 3D scene and to estimate a camera $\mathcal{C}$ that aligns with the input picture, we utilize a geometry transformer model. Those models return an estimated scene camera and back projected 3D points.
We employ a novel strategy to enhance its alignment capabilities: instead of feeding only the original image $\mathbf{I}$, we provide the model with both $\mathbf{I}$ and the "empty room" background image $\mathbf{I}_{bg}$ generated during our inpainting phase (Section \ref{section:novel_inpainting}). The geometric model processes both images and produces two aligned point clouds and corresponding cameras, visualized in our framework Figure~\ref{fig:pipeline}.
We retain only the main camera $\mathcal{C}$ of the input image $\mathbf{I}$, as the second camera is aligned to the background. We also now have the full scene point cloud $\mathcal{P}_{scene}$ and the aligned background point cloud $\mathcal{P}_{bg}$. \\
\textbf{Background Creation.}
We can use the extracted background point cloud $\mathcal{P}_{bg}$ as a base for our background creation. As there can be minor offsets between $\mathcal{P}_{scene}$ and $\mathcal{P}_{bg}$, an iterative algorithm aligns both scenes by slightly shifting them in global 3D space. 
In the end, a meshing algorithm turns the aligned point cloud into a usable mesh for simulations and rendering.\\
\textbf{Object Point Cloud Extraction.}
To create specific 3D targets for our alignment loss, we use the 2D object masks $\mathbf{M}_{obj}$ from the segmentation step (Section \ref{section:image_extraction}). By back projecting these masks into the 3D scene, we effectively stencil $\mathcal{P}_{scene}$ and extract the specific points belonging to each object. This results in a set of object specific point clouds $\{\mathcal{P}_{target}^1, ..., \mathcal{P}_{target}^N\}$ for $N$ objects. These point clouds serve as the ground truth for our 3D geometric loss term $L_{3D}$ during the pose refinement stage. \\
\textbf{Generative 3D Asset Creation.}
Running in parallel to the scene reconstruction, our modular pipeline generates a 3D asset $\mathcal{M}$ for each object. Using the inpainted, isolated object images $\{\mathbf{I}_{obj\_fg}^1, ..., \mathbf{I}_{obj\_fg}^N\}$ from Section \ref{section:novel_inpainting} as input, we pass each image to a 2D to 3D generative model, generating separated and textured objects.

\subsection{Differentiable Scene Reconstruction}
\label{section:scene_recon}

\textbf{Pose Initialization and Model Selection}.
We first perform a coarse initialization of the object's pose. This strategy is contingent on the object's placement, which we determine by computing the 2D Intersection over Union (IoU) between the object mask $\mathbf{M}_{obj}$ and the floor mask $\mathbf{M}_{floor}$.
If there is no intersection, the object is assigned to a 5-DoF Model (optimizing 3D translation, 1D yaw, and 1D scale) and is initialized by aligning its Oriented Bounding Box (OBB) with that of the target point cloud $\mathcal{P}_{target}$.
If the $\text{IoU}$ is greater than zero, we assume the object is on the floor and assign it to our novel 4-DoF 'planar model'. Its pose is initialized by projecting its bottom vertices onto the fitted floor plane $(\mathbf{n}, \mathbf{p}_0)$, which is extracted from $\mathcal{P}_{target}$ using a robust RANSAC algorithm. \\
\textbf{Composite Loss Function}. \label{sec:loss_function} Both models are optimized by minimizing the same composite loss function $L_{total}$, which ensures alignment to both 2D and 3D priors:
\[
L_{total} = w_{sil} L_{silhouette} + w_{3D} L_{3D} + w_{bbox} L_{bbox},
\]
where $w_{sil}$, $w_{3D}$, and $w_{bbox}$ are scalar weights. \\
\textbf{2D Silhouette Loss ($L_{silhouette}$)}. This term enforces 2D alignment by minimizing the discrepancy between the rendered silhouette $\mathbf{P}$ and the ground truth mask $\mathbf{M}_{obj}$. It is a combination of Dice and Focal loss to robustly handle class imbalance at object edges. \\
\textbf{3D Geometric Loss ($L_{3D}$)}. This term enforces 3D structural alignment by minimizing the point to mesh face distance from the target object point cloud $\mathcal{P}_{target}$ to the surface of the transformed object mesh $\mathcal{M}'$. \\
\textbf{Background Bounding Box Loss ($L_{bbox}$)}. This term acts as a physical regularizer, preventing the object from penetrating static scene geometry. It applies a penalty to any object vertex that lies outside a precomputed background bounding box $\mathcal{B}_{bg}$ on the X and Z axes, while ignoring the Y-axis to allow objects to rest on the floor.\\
\textbf{4-DoF Constrained Planar Optimization.} Our primary contribution for physically plausible scene assembly is the PlanarModel. This approach constrains the 5-DoF problem to a 4-DoF problem defined in a \textit{plane local coordinate system} derived from the fitted ground plane $(\mathbf{n}, \mathbf{p}_0)$. The model's four learnable parameters are defined in this local space: 2D translation $(t_x', t_z')$, 1D yaw $(r_y')$, and 1D uniform scale $(s)$. The key physical constraint is enforced by applying the translation as a 3D vector $(t_x', 0, t_z')$ in this local space, explicitly restricting all movement to the 2D floor surface. In each step, the model applies these 4-DoF transformations locally, then projects the resulting vertices $\mathcal{V}_{plane}$ back to world space using a fixed transformation matrix $\mathbf{T}_{plane \to world}$ to be evaluated by the composite loss. This reduction in dimensionality and enforcement of the strong physical planar placement constraint make the optimization highly robust.

\section{Experiments}
In this section, we conduct a comprehensive evaluation of our proposed framework, 3D-RE-GEN, to validate its effectiveness in reconstructing high fidelity, physically plausible 3D scenes from a single image. We first outline our implementation details and the standard metrics used for evaluation. We then present quantitative and qualitative results, comparing our method against SOTA compositional scene reconstruction pipelines. These comparisons demonstrate that our framework achieves SOTA performance, particularly in generating complete scenes with correctly aligned assets and a fully modeled background.

Finally, we perform ablation studies to assess the impact of our novel contributions. These studies isolate the components of our method, specifically validating the effectiveness of our Application-Querying inpainting technique and the crucial role of our 4-DoF constrained PlanarModel in achieving greater physical and spatial alignment.

\begin{figure*}[h]
  \centering

  \includegraphics[width=1\linewidth]{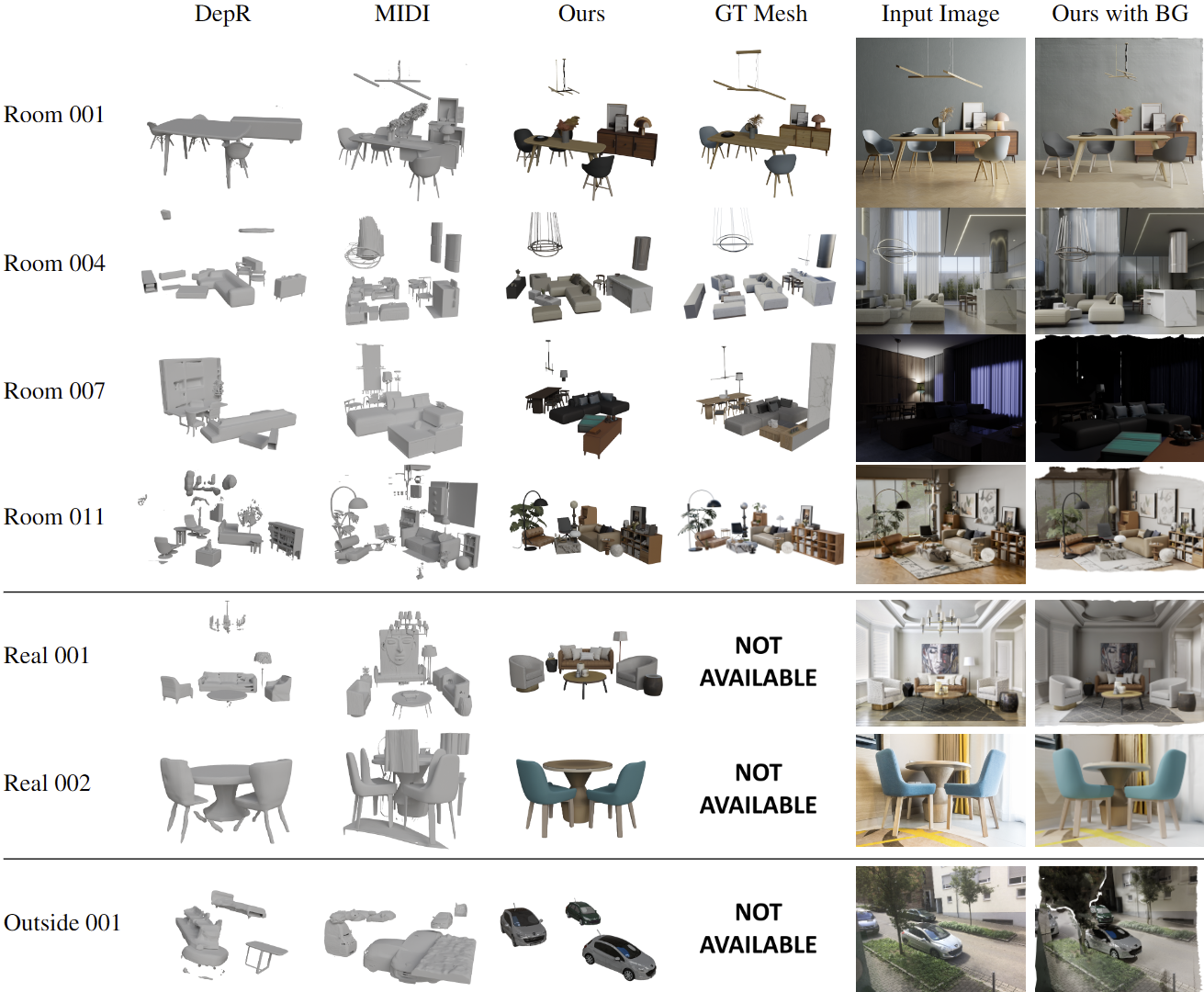}

  \caption{
    Qualitative comparison across different methods
    for different input scenes. Starting with 4 scenes based on synthetic datasets and two real images. In the bottom line, we even tested an outside image.
  }
  \label{fig:qual_comparison}
\end{figure*}

\subsection{Experimental Overview}
\textbf{Implementation Details.}
3D-RE-GEN~integrates several off the shelf models: GroundedSAM \cite{renGroundedSAMAssembling2024} for object detection and segmentation; Googles Image Flash (NanoBanana) \cite{GoogleNanoBanana2025, comaniciGemini25Pushing2025} for image modification and inpainting, Hunyuan3D 2.0 \cite{zhaoHunyuan3D20Scaling2025} for 3D asset reconstruction, and VGGT \cite{wangVGGTVisualGeometry} for camera parameter estimation and point cloud reconstruction.  
Scene reconstruction is performed using a differentiable renderer implemented with PyTorch3D  \cite{raviAccelerating3DDeep2020Pytorch3d} which is used for pose estimation per object based on loss. Blender \cite{Blender} is used as a physically based renderer for visualization and evaluation. Runtime depends on available hardware resources. \\ The system supports multi GPU execution, distributing tasks such as object placement and 2D to 3D asset generation across devices. On a single GPU, a typical scene with approximately 10 objects requires 17 to 20 minutes, with 3D asset creation being the dominant factor. Using four GPUs reduces this to around 7 to 8 minutes. Experiments were conducted on an NVIDIA RTX 4090, 24GB VRAM and an NVIDIA A40, 40GB VRAM. \\ The most memory intensive component is Hunyuan3D 2.0; replacing it with lighter models such as SPAR3D \cite{huangSPAR3DStablePointAware2025} would allow runtime with as 16GB VRAM or less.\\

\textbf{Datasets.} Since 3D-RE-GEN~is a non trained, model based method, it does not require large scale training datasets. Instead, we evaluate it on a set of hand picked scenes synthetic scenes from CGTrader \cite{CGTrader3DModel}, covering diverse conditions such as cluttered rooms, low light scenes, and monochromatic environments.\\ We use royalty free images for our real world scenario cases and some self taken pictures for outside scene tests.
These manually selected scenes provide strong generalization tests, as many modern approaches, such as MIDI \cite{huangMIDIMultiInstanceDiffusion2024}, DepR \cite{zhaoDepRDepthGuided2025}, and SceneGen \cite{mengSceneGenSingleImage3D2025}, are trained on large synthetic datasets like 3D-Front \cite{fu3DFRONT3DFurnished2021}. Using unseen and varied CGTrader scenes helps ensure that evaluation metrics reflect realistic generalization rather than dataset specific bias.\\

\textbf{Metrics.} All 3D metrics are computed at the scene level. For evaluation, both the predicted and ground truth scenes are converted into point clouds and normalized to a unit scale centered at the world origin \([0, 0, 0]\). The two point clouds are aligned using the ICP algorithm \cite{Besl_1992_ICP} before metric computation.  
Quantitative 3D evaluation includes commonly used metrics like Chamfer Distance (CD), F-score and Bounding Box IoU (BBOX-IOU) \cite{huangMIDIMultiInstanceDiffusion2024, zhaoDepRDepthGuided2025, mengSceneGenSingleImage3D2025, zhouZeroShotSceneReconstruction2024}. Precision as part of F-Score allows for a more intricate overview. BBOX-IOU as a metric highlights if the objects are actually at the same relative scene position and size. The Hausdorff distance~\cite{alibekovAdvancingPrecisionMultiPoint2025} was chosen to show the behavior of outliers, highlighting more consistent scenes that are spatially cohesive.\\
\textbf{Baselines.} 
We compare 3D-RE-GEN~against two SOTA scene generation methods, MIDI \cite{huangMIDIMultiInstanceDiffusion2024} and DepR \cite{zhaoDepRDepthGuided2025}. Both are evaluated using their publicly available pretrained models. As both rely on segmentation masks as input, we utilize their respective automatic mask generation modules, each based on GroundedSAM derivatives.  
While MIDI produces textured outputs, DepR generates geometry only. As texture generation in MIDI currently faces known issues of reproducibility and doesn't contribute to 3D metrics we omit it.\\

\begin{table}[h]
    \centering
    \begin{tabular}{lccc}
        \toprule
        Metric 3D & DepR \cite{zhaoDepRDepthGuided2025} & MIDI \cite{huangMIDIMultiInstanceDiffusion2024} & Ours \\
        \midrule
        CD~$\downarrow$ & 0.028 & 0.036 & \textbf{0.011} \\ \addlinespace
        F-Score~$\uparrow$ & 0.65 & 0.70 & \textbf{0.85} \\ \addlinespace
        IOU~$\uparrow$ & 0.44 & 0.57 & \textbf{0.63} \\ \addlinespace
        Precision~$\uparrow$ & 0.12 & 0.15 & \textbf{0.21} \\ \addlinespace
        Recall~$\uparrow$ & 0.18 & 0.13 & \textbf{0.19} \\ \addlinespace
        Hausdorff~$\downarrow$ & 0.61 & 0.55 & \textbf{0.33} \\
        \bottomrule
    \end{tabular}
    \caption{Quantitative comparison of different methods tested on our synthetic collected dataset, see implementation details.}
    \label{tab:quant_metrics}
\end{table}

\subsection{Results}

\textbf{Quantitative Results.} Table~\ref{tab:quant_metrics} summarizes the quantitative evaluation on synthetic data of our final scene reconstructions. 3D-RE-GEN~consistently and significantly outperforms competing methods such as MIDI and DepR across all major 3D metrics. While MIDI achieves a respectable BBOX-IOU due to its multi instance attention, our method surpasses it by enforcing a strict, physically based ground alignment. Furthermore, the significantly lower Hausdorff distance of our method indicates a more stable reconstruction quality, with fewer outlier points and artifacts compared to all other approaches.

\begin{figure}[h]
    \centering
        \begin{subfigure}{0.9\linewidth}
        \centering
        \includegraphics[width=\linewidth]{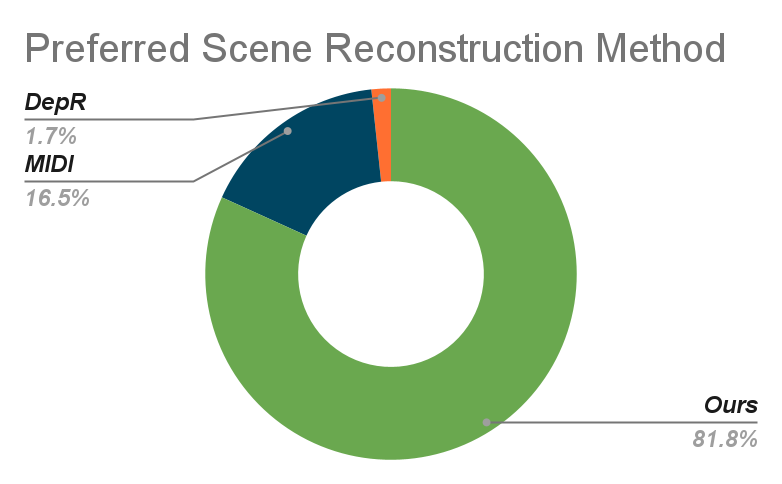}
        \caption{Overall reception}
        \end{subfigure}
    \hfill
        \begin{subfigure}{0.9\linewidth}
        \centering
        \includegraphics[width=\linewidth]{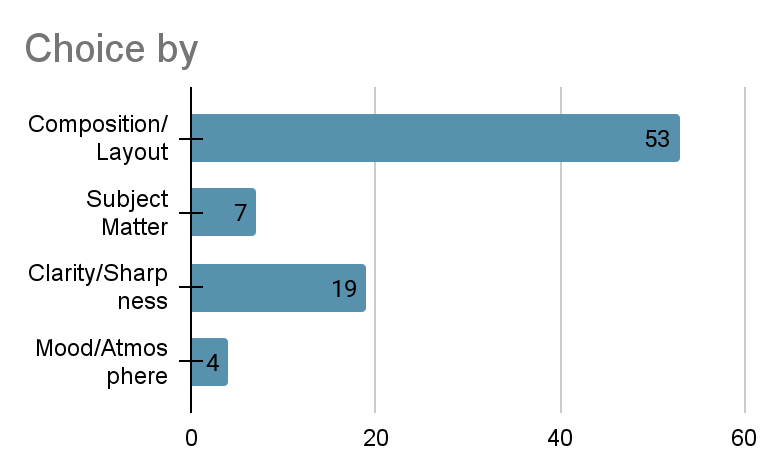}
        \caption{Reason for choice}
        \label{fig:abl_wo_aq}
    \end{subfigure}    
    \caption{Quantitative survey with 59 participants on how people perceive scene reconstruction method outputs.}
    \label{fig:survey_reception}
\end{figure}
\newpage
\textbf{Qualitative Results.} Figure~\ref{fig:qual_comparison} illustrates representative qualitative comparisons, showcasing our method's robust generalization across diverse data, including synthetic, challenging real world, and even outdoor images; a domain rarely tested by existing methods. 3D-RE-GEN~uniquely produces coherent, physically plausible scenes. We note that for real world inputs, our method not only maintains correct spatial grounding and perspective alignment, but also achieves exceptionally high visual fidelity and texture quality.
MIDI, while capturing spatial grouping, frequently exhibits severe mesh artifacts, such as merging distinct objects or duplicating geometry. DepR often fails to produce visually coherent scenes; its outputs, while scoring well on some metrics, often appear as misaligned, incorrectly rotated, or incomplete "blobs" that float in space. In contrast, our method consistently generates well structured layouts with sharply defined, complete geometry and a fully reconstructed, textured background.

The high quality background and the precise alignment of our assets are critical for our target applications, as they provide a complete environment suitable for downstream VFX workflows, such as shadow casting, realistic light bounces, and physics based simulations. Our intermediate results also highlight the strength of our pipeline. Our Application-Querying method produces clean, complete object assets even from heavily occluded inputs. Figure \ref{fig:qual_comparison} further shows our background reconstruction, which provides the essential physically grounded stage upon which our 4-DoF alignment can produce a final, cohesive scene.

We conducted a small study with 59 participants \ref{fig:survey_reception}. They had to chose, based on an input image, which output of a scene reconstruction model they preferred most. Visually it resembled closely our qualitative table \ref{fig:qual_comparison}. 3D-RE-GEN achieved the best score of 81\% of reception of quality. The most picked reason of choice was : "Layout / Composition".

\begin{figure}[h]
    \centering
        \begin{subfigure}{0.48\linewidth}
        \centering
        \includegraphics[width=\linewidth]{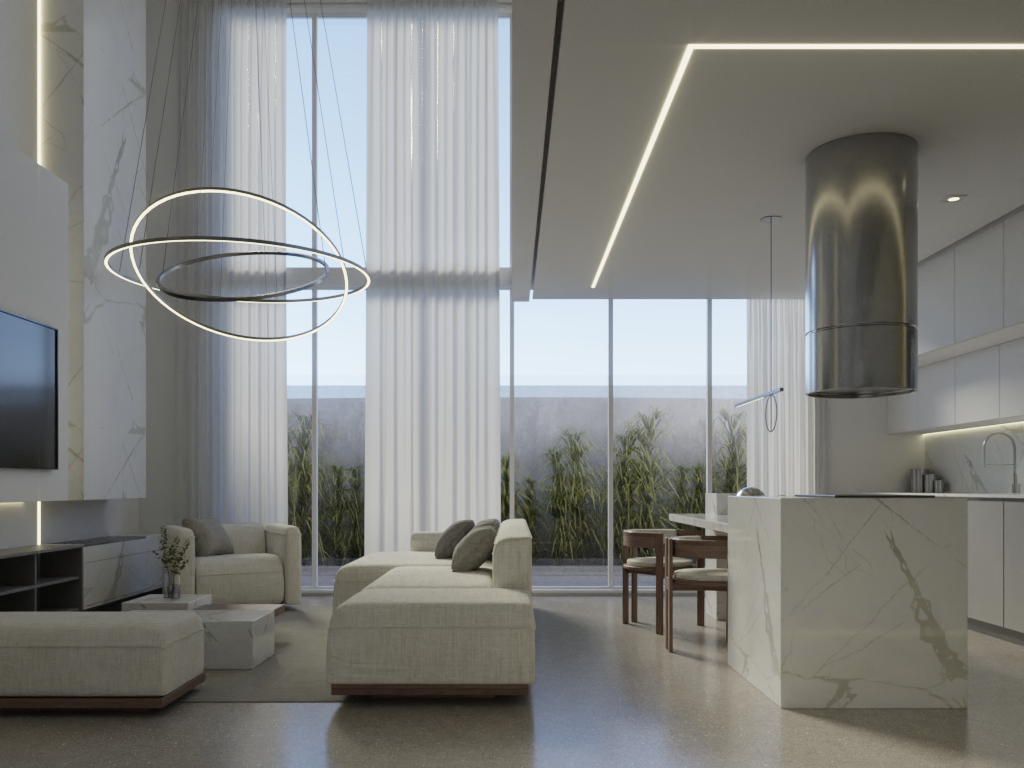}
        \caption{Input image}
        \label{fig:abl_wo_aq}
    \end{subfigure}
    \hfill
        \begin{subfigure}{0.48\linewidth}
        \centering
        \includegraphics[width=\linewidth]{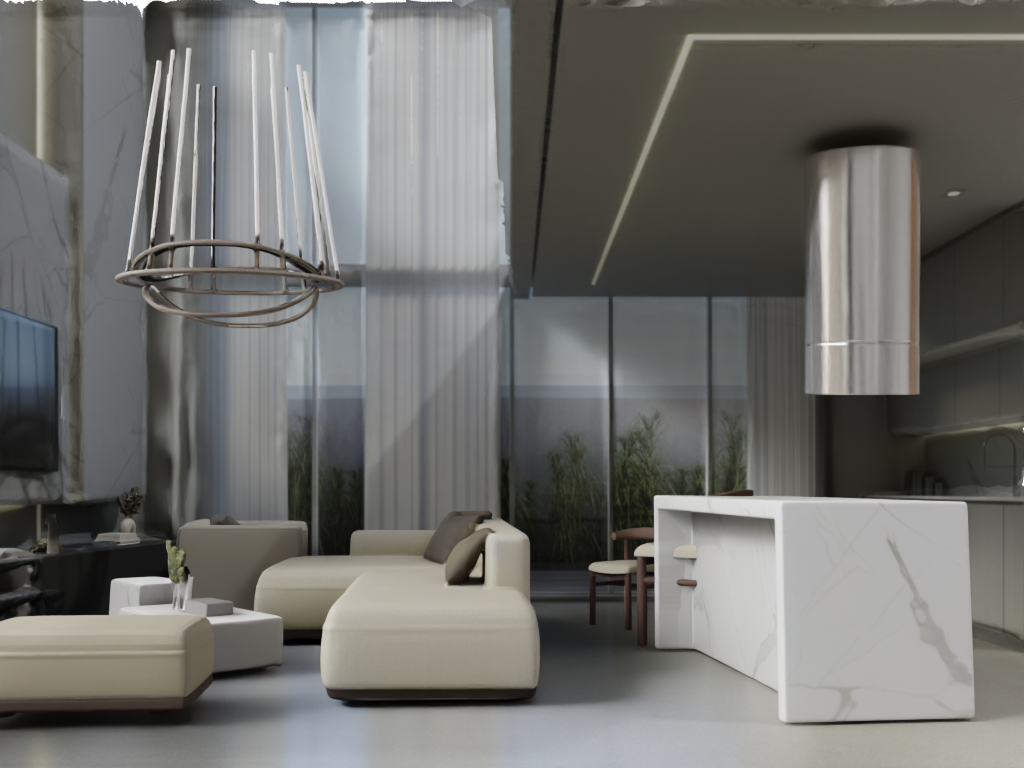}
        \caption{3D-RE-GEN~output}
        \label{fig:abl_wo_aq}
    \end{subfigure}
    \hfill
    \begin{subfigure}{0.48\linewidth}
        \centering
        \includegraphics[width=\linewidth]{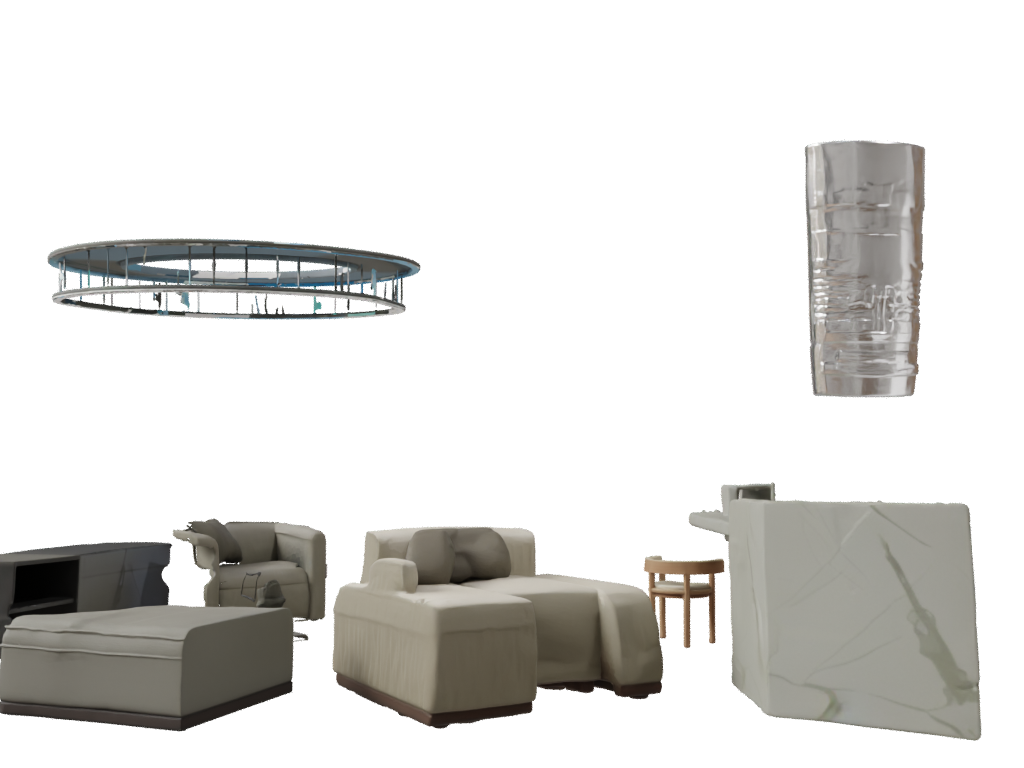}
        \caption{Without A-Q}
        \label{fig:abl_wo_aq}
    \end{subfigure}
    \hfill
    \begin{subfigure}{0.48\linewidth}
        \centering
        \includegraphics[width=\linewidth]{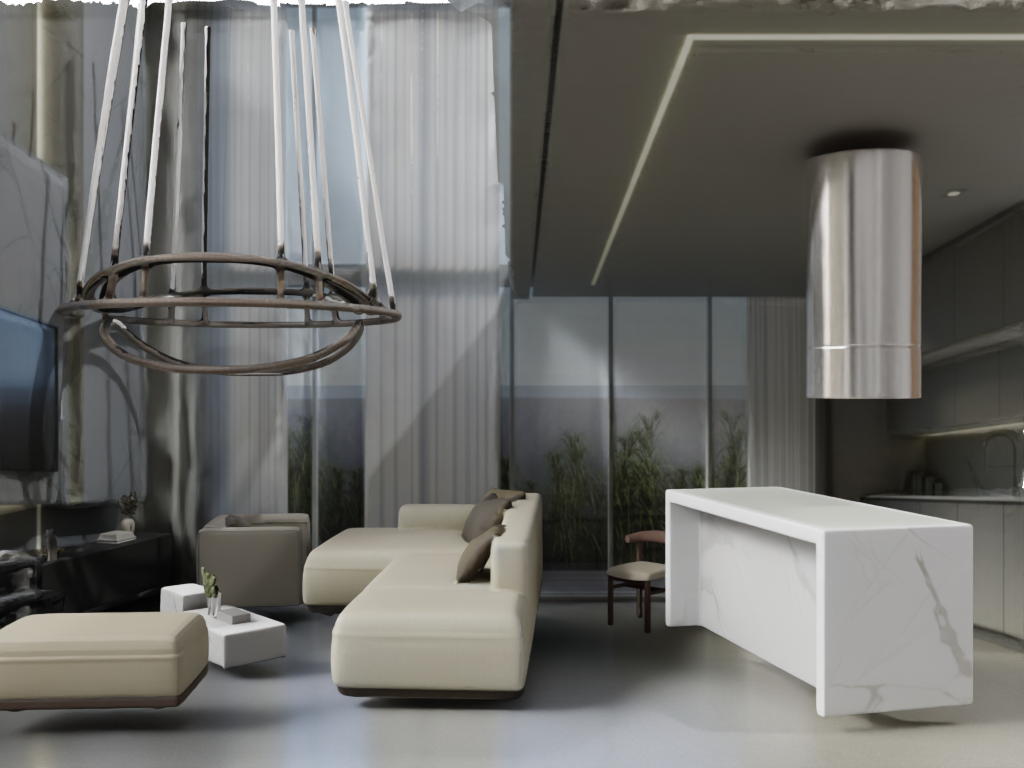}
        \caption{Without 4-DoF}
        \label{fig:abl_wo_4dof}
    \end{subfigure}

    \caption{\textbf{Ablation examples.} Output of 3D-RE-GEN without the use of our Application-Querying Model \ref{fig:abl_wo_aq} and without the use of our ground constrain 4-DoF Model \ref{fig:abl_wo_4dof}.}
    \label{fig:ablation_tests}
\end{figure}

\begin{table}[h]
    \centering
    \begin{tabular}{lccc}
        \toprule
        Metric 3D & No 4-DoF & No A-Q  & Ours \\
        \midrule
        CD~$\downarrow$         & 0.019 & 0.030 & \textbf{0.011} \\ \addlinespace
        F-Score~$\uparrow$      & 0.82 & 0.68 & \textbf{0.85} \\ \addlinespace
        IOU ~$\uparrow$         & 0.52 & 0.51 & \textbf{0.63} \\ \addlinespace
        Hausdorff ~$\downarrow$ & 0.51 & 0.52 & \textbf{0.33} \\
        
        \midrule
        Metric 2D & No 4-DoF & No A-Q  & Ours \\
        \midrule
        SSIM ~$\uparrow$        & 0.27 & 0.13 & \textbf{0.29} \\ \addlinespace
        LPIPS ~$\downarrow$       & 0.46 & 0.66 & \textbf{0.43} \\ \addlinespace
        \bottomrule
    \end{tabular}
    \caption{\textbf{Ablation study.} Comparing the complete 3D-RE-GEN model against 3D-RE-GEN without ground constraints (No 4-DoF); and against 3D-RE-GEN without Application-Querying model (No A-Q).}
    \label{tab:quant_metrics_abl}
\end{table}

\subsection{Ablation Study}
\label{sec:ablation}
To validate the effectiveness of our key contributions, we conduct an ablation study, with quantitative results presented in Table \ref{tab:quant_metrics_abl} and visually in Figure \ref{fig:ablation_tests}. We evaluate our full pipeline against two modified versions. Firstly, 3D-RE-GEN without the use of our 4-DoF model, so pose estimation without ground constraints. All objects are pose optimized using only the 5-DoF RegularModel. This tests how much objects would usually deviate from following camera perspective and how much our 4-DoF model ensures a physically aligned scene creation. Secondly, 3D-RE-GEN without A-Q: This version disables our Application-Querying inpainting step. The 2D to 3D generator model is instead fed with only the unrepaired segmented object.

We evaluate these versions using both 3D geometric metrics and 2D perceptual metrics (SSIM and LPIPS \cite{huangMIDIMultiInstanceDiffusion2024, mengSceneGenSingleImage3D2025}) to measure the visual quality of the final rendered image against the input. Our full, combined model shows the best performance, achieving the best scores across all metrics. The model without 4-DoF PlanarModel suffers from drops in 3D metrics, proving that our 4-DoF ground plane constraint is critical for geometric accuracy and well structured scenes follow camera perspective. Without the use of the 4-DoF ground constraints, objects can float and don't need to follow correct ground alignment, see figure \ref{fig:abl_wo_4dof}. The version without Application-Querying \ref{fig:abl_wo_aq} shows a strong falloff  in both 2D and 3D scores, demonstrating that generating assets from incomplete, occluded segments leads to poor geometry and visual incoherence and no background creation. Objects are recreated in 3D like the segmented input directs, so no scene context is given and incomplete objects are created. These ablations confirm that our pipeline's SOTA performance is a direct result of our novel 4-DoF optimization and our context aware Application-Querying.

\section{Conclusion}
\label{sec:Conclusion}

In this work, we introduced \paperName, a novel and robust framework for compositional reconstruction of complete 3D indoor scenes from a single input image. Our method leverages large scale models to decompose an image, while two key novelties address the most significant challenges in scene reconstruction. First, our context aware \textbf{Application-Querying (A-Q)} inpainting technique allows generation of high quality and complete 3D assets, even from occluded inputs. Our \textbf{4-DoF PlanarModel} ensures physical plausibility by correctly aligning all ground based objects to a interpolated floor plane. Our pipeline successfully recovers the background, a matched camera, and all generated objects, ensuring structural coherence and correct perspective alignment with the input view. Quantitative and qualitative evaluations demonstrate that \paperName~achieves SOTA performance, producing cohesive 3D scenes free of the floating artifacts and geometric inconsistencies common in other methods. The reconstructed scenes, from a single image, are readily applicable to downstream VFX and game development workflows, offering an efficient and powerful tool for easy usage.

\begin{section}{Acknowledgements}
    \label{sec:acknowledgements}

    Funded by the Deutsche Forschungsgemeinschaft (DFG, German Research Foundation) under Germany's Excellence Strategy – EXC number 2064/1 – Project number 390727645.
    This work was supported by the German Research Foundation (DFG): SFB 1233, Robust Vision: Inference Principles and Neural Mechanisms, TP 02, project number: 276693517.
    This work was supported by the Tübingen AI Center.
    The authors thank the International Max Planck Research School for Intelligent Systems (IMPRS-IS) for supporting Jan-Niklas Dihlmann.
\end{section}

{
    \small
    \bibliographystyle{ieeenat_fullname}
    \bibliography{main}
}
\newpage
\appendix
\clearpage
\setcounter{page}{1}
\maketitlesupplementary

\section*{Overview}
This supplementary material complements the main manuscript by providing extra background creation specifics, extended qualitative results, and a critical discussion of the proposed 3D-RE-GEN framework. The document is organized as follows:

\begin{itemize}
    \item \textbf{Additional Videos (Section \ref{suppl:add_vid}):} We introduce the accompanying video materials that demonstrate the pipeline's efficacy and downstream utility in dynamic environments.
    \item \textbf{Discussion (Section \ref{sec:discussion_suppl}):} We provide a critical analysis of the system's current limitations and outline prospective avenues for future research and development.
    \item \textbf{Background Extraction and Texturing (Section \ref{sec:background_suppl}):} We offer a detailed visualization of the workflow used to extract geometry and generate textures for the scene background.
    \item \textbf{Prompt Specifications (Section \ref{sec:prompts}):} To ensure reproducibility, we provide the specific text and image prompts utilized in our Application-Querying module.
    \item \textbf{User-Friendly Mask Refinement (Section \ref{suppl:mask_gen}):} We demonstrate our custom Gradio interface, designed to facilitate the manual fine tuning of segmentation masks for enhanced reconstruction precision.
\end{itemize}

\section{Additional Videos}
\label{suppl:add_vid}
We provide two supplementary video files to visually validate the capabilities of our method:

\textbf{Downstream Application.} 
\texttt{3D-RE-GEN\_vfx.mp4} demonstrates the practical utility of our reconstructed scenes in professional workflows. This sequence showcases a visual effect applied directly to our generated output, highlighting the mesh quality and physical plausibility necessary for seamless integration into VFX pipelines.

\textbf{Method Overview and Comparisons.} 
\texttt{3D-RE-GEN\_overview\_and\_results.mp4} provides a comprehensive summary of the framework. It presents qualitative comparisons against state of the art methods using 360 degree turntable visualizations, allowing for a rigorous assessment of geometric fidelity and texture consistency.

\section{Discussion}
\label{sec:discussion_suppl}
In this section we highlight some issues that can occour or overall weaknesses. In the end we show what we plan on improvements that others can explore or we will implement in the future.

\subsection{Limitations}
\label{subsec:limitations}

\textbf{Mask Sensitivity.} The proposed method exhibits sensitivity to inaccuracies in the initial segmentation masks, which are fundamental for guiding both inpainting and spatial positioning. Segmentation errors can propagate through the pipeline, providing erroneous context to the Application Querying module \cite{GoogleNanoBanana2025}, resulting in the generation of incorrect geometry. Furthermore, inaccurate masks may cause the inadvertent culling of valid points from the target point cloud used for the 3D loss. This degradation of the optimization target can lead to suboptimal convergence, as the resulting 3D geometry fails to align with the intended 2D silhouette constraints.

\textbf{Geometric Estimation Uncertainty.} The probabilistic nature of the geometry transformer introduces potential reconstruction artifacts. Points classified with low confidence are discarded during preprocessing, which can compromise the structural integrity of the reconstructed scene, manifesting as holes or discontinuities in the background mesh (see Figure \ref{fig:suppl_bg_extract}). Additionally, because the geometry transformer is not explicitly trained to align background only images with original scene images under a unified coordinate system, minor discrepancies in camera estimation can occur, potentially resulting in spatial misalignment between the generated background and the global scene context.

\textbf{Optimization Convergence.} As with many learning based optimization frameworks, the differentiable rendering process is susceptible to the non  convex nature of the loss optimizations. The pose refinement is liable to get trapped in local minima, particularly when the initial pose estimate deviates significantly from the ground truth. Silhouettes for example can be very close to the current perspective of an objects which is rotated 180 degrees. This can occasionally cause objects to converge to suboptimal positions or orientations within the scene.

\textbf{Object Granularity.} To balance computational efficiency with reconstruction robustness, complex compound objects are currently generated as unified meshes rather than discrete assemblies (e.g., a bookshelf and its books are treated as a single entity). While this simplifies the reconstruction process, it limits the granularity of downstream interactions in VFX workflows, potentially necessitating manual geometry separation for fine  grained manipulation.

\textbf{Generative Stochasticity.} A final limitation arises from the inherent stochastic nature of the generative components. The reliance on random seeds for both context  aware inpainting and 2D to 3D asset creation means that identical inputs can yield perceptually distinct results, posing a challenge for applications requiring strictly deterministic reproducibility.

\subsection{Future Work}
\label{subsec:future_work}
The modular architecture of \paperName opens several promising avenues for future research and functional expansion.

\textbf{Hierarchical and Non  Planar Constraints.} While our current 4-DoF optimization strictly enforces ground plane alignment, future work will aim to generalize this to a hierarchical constraint system. This would enable the placement of "sub objects" onto arbitrary planar surfaces, such as placing lamps on tables or books on shelves, facilitating the decomposition of clustered assets into granular, interactive components.

\textbf{Multi View.} Although designed for single image reconstruction, the optimization framework is extensible to multi view inputs. Integrating constraints from multiple viewpoints would significantly reduce geometric ambiguity and improve occluded region fidelity. 

\textbf{Advanced Rendering and Simulation.} To further enhance utility for VFX pipelines, we plan to integrate advanced material estimation models capable of outputting high fidelity BRDF parameters. This would allow for photorealistic relighting and seamless integration of reconstructed scenes into varying lighting environments.

\textbf{Generalization to Unconstrained Environments.} Finally, we demonstrate the extensibility of our framework beyond indoor scenes. By adapting our ground alignment logic to accommodate uneven terrain  and leveraging the robustness of our 4-DoF optimization, \paperName~ successfully reconstructs complex, unconstrained outdoor environments (see Figure \ref{fig:qual_comparison}). We observe that our pipeline correctly extracts and aligns structured outdoor assets, such as vehicles, to the estimated ground plane. However, we note a current limitation in the generative 2D to 3D backend: models trained primarily on synthetic or object centric datasets \cite{zhaoHunyuan3D20Scaling2025} struggle with the abstract, high frequency geometry of organic assets like trees and foliage. While our spatial positioning remains accurate, the geometric fidelity of these specific asset classes represents a domain gap to be bridged by future generative models trained on diverse outdoor data.

\section{Background Extraction and Texturing}
\label{sec:background_suppl}
\begin{figure}[h]
    \centering
    \includegraphics[width=1\linewidth]{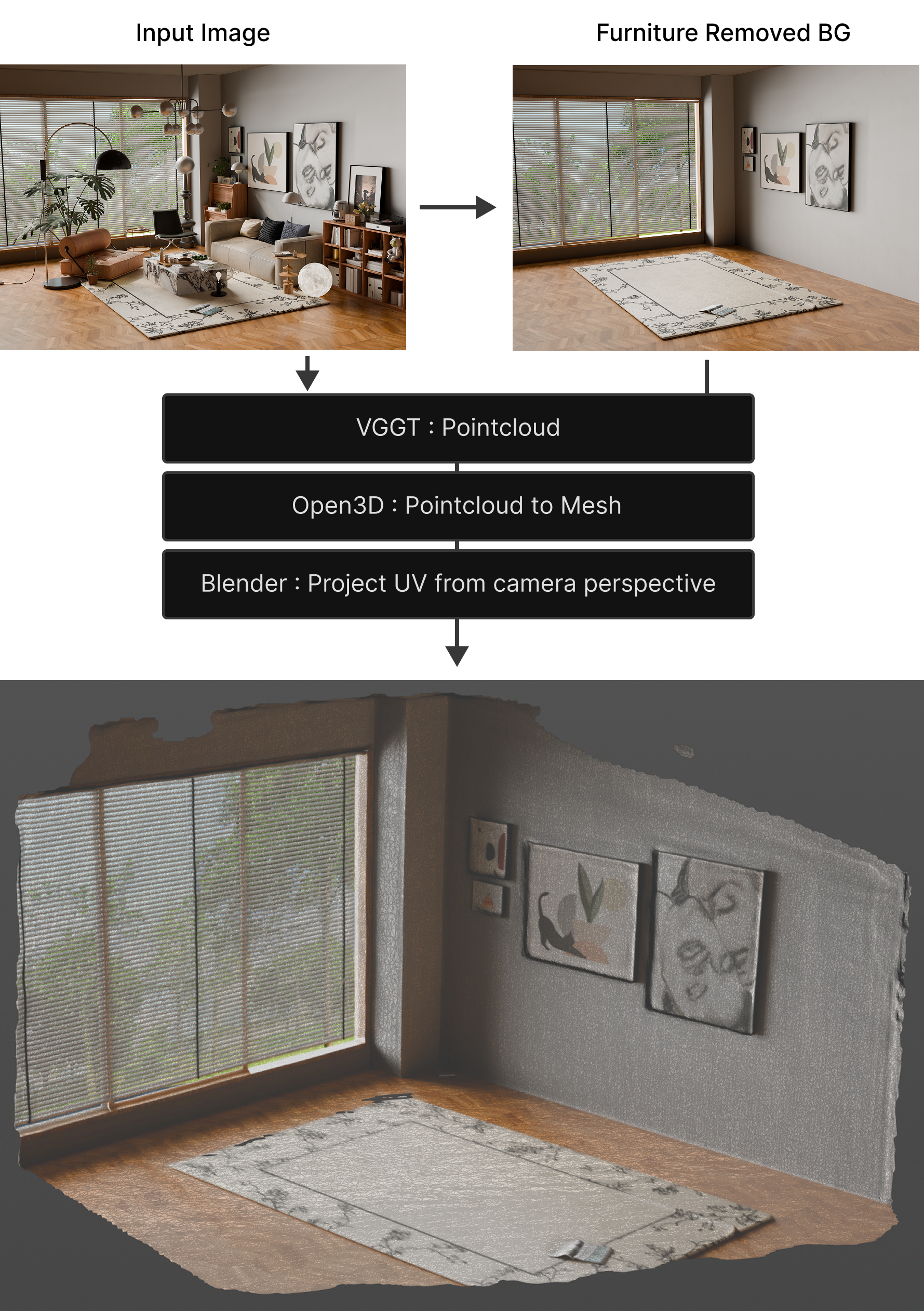}
    \caption{\textbf{Background extraction workflow.} Extraction of pointclouds trough geometry transformer; pointclouds getting meshed with surfacing algorithm and then texture applied through projection.}
    \label{fig:suppl_bg_extract}
\end{figure}

The reconstruction of the background environment is still an important part of our pipeline, ensuring a spatially correct stage for object placement (BBOX-Loss) and an object to bounce lights and act as collision for simulations. Due to space constraints in the main manuscript, we provide a more detailed workflow for background creation here, as illustrated in Figure \ref{fig:suppl_bg_extract}.

\textbf{Geometry Extraction.}
We use VGGT \cite{wangVGGTVisualGeometry} to extract the initial point cloud representation of the scene. A key strategy in our approach is to process both the original input image and the inpainted "empty room" (background only) image simultaneously. This enforces a shared coordinate system, ensuring the extracted background geometry is perfectly aligned with the original camera perspective.

\textbf{Meshing.}
To facilitate standard VFX workflows, we convert the raw point cloud into a clean triangle mesh. We utilize the Poisson surface reconstruction algorithm implementation in Open3D \cite{zhouOpen3DModernLibrary2018} to generate a surface from the point cloud data.

\textbf{Texturing and Material Generation.}
For texturing, the mesh is imported into Blender \cite{Blender}, where we perform a camera based UV projection. We project the inpainted "empty room" image directly onto the geometry, effectively using it as a baked light texture. While we also generate a full suite of PBR maps including Albedo, Roughness, Metallic, and Normals; using Marigold \cite{keMarigoldAffordableAdaptation2025}. We  observed that the original baked light image yielded better visual quality and coherence for the background compared to the decoupled PBR maps.

\section{Prompts}
\label{sec:prompts}
This chapter gives a insight in how we utilized text and image prompts to guide the models to the best result.

\textbf{Prompts for GroundedSAM. }  
Starting point for image segmentation is GroundedSAM \cite{renGroundedSAMAssembling2024}. We use diverse text prompts to cover the largest base, especially for bigger furniture. Most used prompts are: \\
"
  -furniture 
  -table with decorations 
  -chair 
  -sideboard with decorations 
  -shelf 
  -bookshelf 
  -dresser with decorations 
  -cabinet with decorations 
  -couch 
  -bed with pillows 
  -lamp 
  -floor 
  -kitchen counter
"

\begin{figure}[h!]
    \centering
    \includegraphics[width=0.95\linewidth]{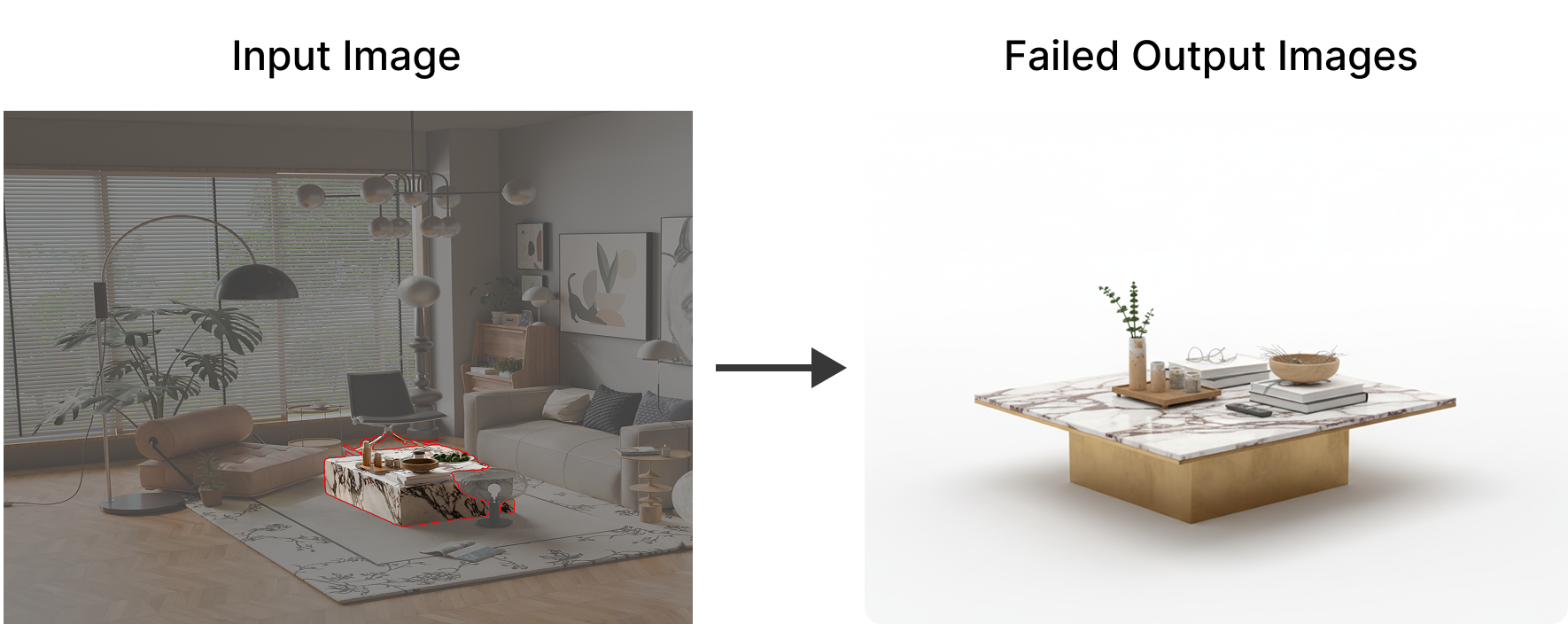}
    \caption{Using just an inline approach can lead to wrong scene understanding.}
    \label{fig:inline_prompt}
\end{figure}

\textbf{Prompts for nanoBanana.} 
The inputs for nanoBanana \cite{GoogleNanoBanana2025} needed a bit more refinement with a lot of trail and error over time. Following are the prompts that worked overall the best.

First was a semi working \textbf{inline prompt} visualized in figure \ref{fig:inline_prompt}. It was a starting point but some cases didn't work correctly. For example issues were that a wrong object was picked, a wrong object understanding, nothing at all happens (gave back the same image as the input was) and the inline color was mistakenly seen as part of the object : \\ 
"
Extract this red marked {object}.
Create a single render of it with a white background.
" \\

\textbf{Application-Querying prompt}: \\
"
[OBJECT EXTRACTION APPLICATION]: extract a single 3D object out of a scene. The extracted object should appear in the white box without background from a frontal view. Only the single selected object with border should be extracted and repaired if parts are missing. No object occluding the selected object should be reconstructed. No accidental background leaking should be included. Use the scene as a reference and extract the object. 
" \\ 

For the background, we used a prompt with a bit more details. Some backgrounds need more iterations with a different seed to get the best possible outcome. The \textbf{background prompt} with the best working results was as follows: \\
"
Remove ALL objects and furniture. I want a single empty room. No chairs, tables, lamps, dresser, kitchen parts etc.
Just give me back the same room but EMPTY. Keep only canvas and rugs. Same light, same perspective, same walls, floor and ceiling.
"

\begin{figure}[h!]
    \centering
    \includegraphics[width=1\linewidth]{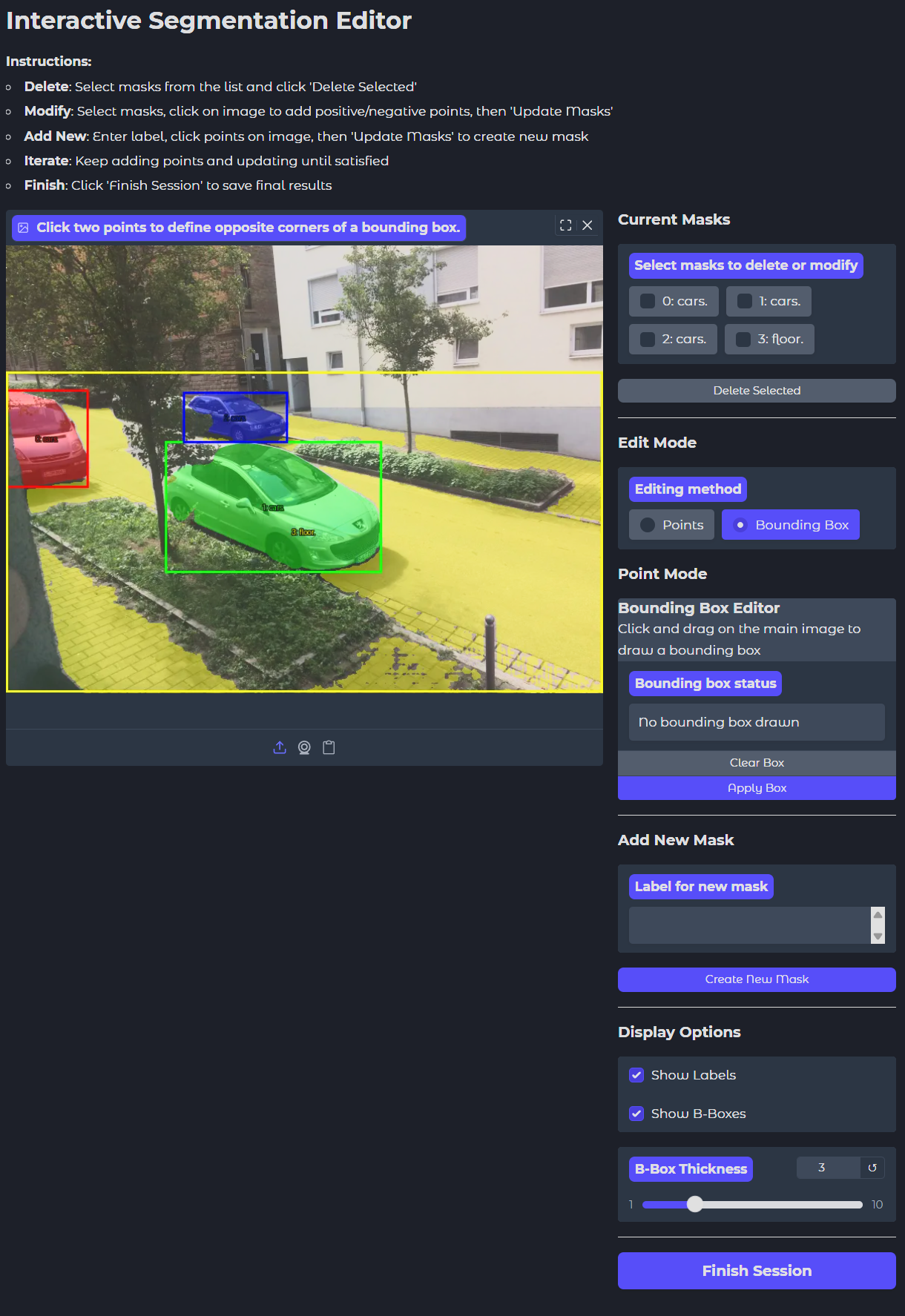}
    \caption{Gradio App to finetune masks from GroundedSAM}
    \label{fig:gradio_app}
\end{figure}

\section{User-friendly mask generation}
\label{suppl:mask_gen}

Mask generation is still a hard topic. Latest models like SAM \cite{kirillovSegmentAnything2023} still struggle to correctly classify objects consistently. Research tries to negate the issues by improving with extra prompts \cite{daiSAMAugPointPrompt2024} or trying to improve with better training \cite{quDeOcc1to33DDeOcclusion2025}. We used SAM-HQ \cite{keSegmentAnythingHigh2023} for better mask initialization, which improved quality drastically. Even still, some objects where not covered good enough. As masks are fundamental for a good result, we created a gradio \cite{abidGradioHassleFreeSharing2019}
app that lets the user add, manipulate and delete masks. For usage a gradio UI can be opened up in a browser locally \ref{fig:gradio_app}.

\end{document}